\documentclass[10pt,twocolumn,letterpaper]{article}

\usepackage{cvpr}              % To produce the CAMERA-READY version
\usepackage{graphicx}
\usepackage{amsmath}
\usepackage{amssymb}
\usepackage{booktabs}
\usepackage{multirow}
\usepackage{xcolor}
\usepackage[accsupp]{axessibility}
\usepackage{url}

\usepackage[pagebackref,breaklinks,colorlinks]{hyperref}

\usepackage[capitalize]{cleveref}
\crefname{section}{Sec.}{Secs.}
\Crefname{section}{Section}{Sections}
\Crefname{table}{Table}{Tables}
\crefname{table}{Tab.}{Tabs.}

\def\cvprPaperID{7747} % *** Enter the CVPR Paper ID here
\def\confName{CVPR}
\def\confYear{2023}

\newcommand{\lyy}{\color{black}}
\newcommand{\cz}{\color{black}}

\begin{document}
	% \linenumbers
	%\pagestyle{headings}
	%\mainmatter
	%\def\ECCVSubNumber{5633}  % Insert your submission number here
	
	\title{Pose-disentangled Contrastive Learning for Self-supervised Facial Representation 
		
		%Pose-disentangled Contrastive Facial Representation Learning
		
	} % Replace with your title
	\author{Yuanyuan Liu$^{1}$\thanks{Equally-contributed first authors} , Wenbin Wang$^{1*}$, Yibing Zhan$^2$, Shaoze Feng$^1$, Kejun Liu$^1$, Zhe Chen$^3$\thanks{Corresponding author}\\
		$^1$School of Computer Science, China University of Geosciences, Wuhan, China\\
		$^2$JD Explore Academy, China\\
		$^3$The University of Sydney, Australia\\
		%Wuhan, China\\
		{\tt\small \{liuyy, wangwenbin, fengshaoze, liukejun\}@cug.edu.cn; zhanyibing@jd.com; zhe.chen1@sydney.edu.au}
	}
	
	%\begin{comment}
	% \titlerunning{ECCV-22 submission ID \ECCVSubNumber} 
	% \authorrunning{ECCV-22 submission ID \ECCVSubNumber} 
	% \author{Anonymous ECCV submission}
	% \institute{Paper ID \ECCVSubNumber}
	%\end{comment}
	%******************
	\maketitle
	
	\begin{abstract}

		Self-supervised facial representation has recently attracted increasing attention due to its ability to perform face understanding without relying  on large-scale annotated datasets heavily. However, analytically, current contrastive-based self-supervised learning (SSL) still performs unsatisfactorily for {\cz learning} facial representation. {\cz More specifically, existing contrastive learning (CL) tends to learn pose-invariant features that cannot depict the pose details of faces}, {\lyy compromising the learning performance.} 
		To conquer the above limitation of CL, we propose a novel Pose-disentangled Contrastive Learning (PCL) method for {\lyy general} self-supervised facial representation. Our PCL {\cz first devises} a pose-disentangled decoder (PDD) with a delicately designed orthogonalizing regulation, which disentangles the pose-related features from the face-aware features; therefore, pose-related and other pose-unrelated facial information could be {\cz performed} in individual subnetworks and do not affect each other's training. 
		{\cz Furthermore, we introduce a pose-related contrastive learning scheme that learns pose-related information based on data augmentation of the same image, which would deliver more effective face-aware representation for various downstream tasks.} 
		%We conducted a comprehensive linear evaluation on four challenging downstream facial understanding tasks, \textit{ie}, facial expression recognition, face recognition, AU detection and head pose estimation. 
		We conducted linear evaluation on four challenging downstream facial understanding tasks, \textit{ie}, facial expression recognition, face recognition, AU detection and head pose estimation.
		Experimental results demonstrate that our method significantly outperforms state-of-the-art SSL methods.
		%Experimental results demonstrate that our method outperforms cutting-edge contrastive and other self-supervised learning methods with a great margin. 
		Code is available at \href{https://github.com/DreamMr/PCL}{https://github.com/DreamMr/PCL}
		
	\end{abstract}
	%\vspace{-0.3cm}
	
	\section{Introduction}
	Human face perception and understanding is an important and long-lasting topic in computer vision. By analyzing faces, we can obtain various kinds of information, including identities, emotions, and gestures. {\cz Recently, deep convolutional neural networks (DCNNs) \cite{knyazev2017convolutional,gamble2020convolutional,zhao2016deep} have achieved promising facial understanding results, but they require a large amount of annotated data for model training. Since labeling face data is generally a labor- and time-costly process \cite{zhao2018learning}, it becomes important to enable DCNN models to learn from unlabelled face images, which are much easier to collect. Accordingly, researchers have introduced self-supervised learning (SSL) schemes to achieve better learning performance on unlabeled facial data. }
	
	\begin{figure}[t]
		\begin{center}
			\includegraphics[width=1.0\linewidth]{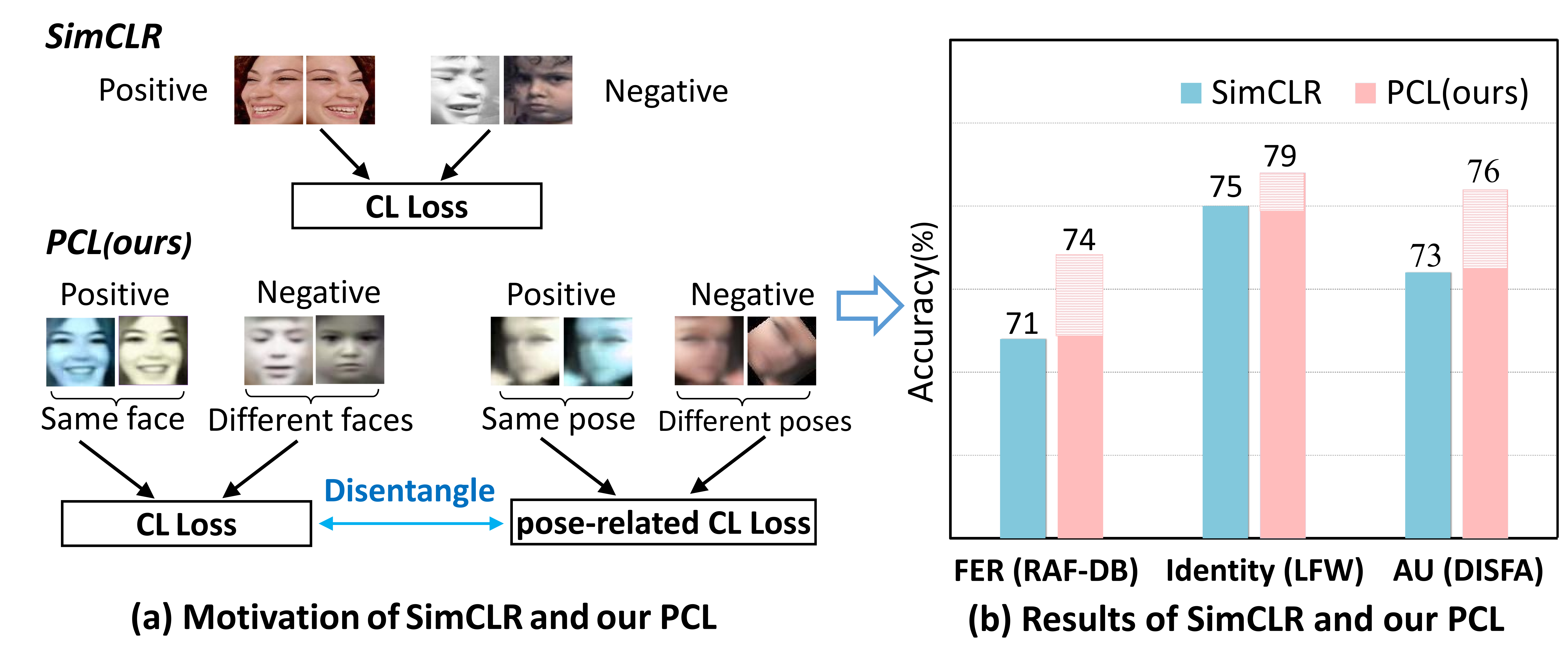}
		\end{center}
		\vspace{-0.3cm}
		\caption{The motivation of our method. Affected by different poses,
			the popular CL method, \eg, SimCLR, treats pose and other face information uniformly, resulting in sub-optimal results.
			%, weakening the learning on facial representation for different down-stream tasks. 
			To alleviate this limitation for CL, our PCL attempts to disentangle the learning on pose-related features and pose-unrelated facial features, thus achieving more effective self-supervised facial representation learning for downstream facial tasks.		
		}
		\label{fig:motivation}
		\vspace{-0.5cm}
	\end{figure}

	{\cz To achieve effective SSL performance, contrastive learning (CL) based strategy is widely applied in the community \cite{mrn2019learning,he2020momentum,chen2020simple}. In general, a CL-based method pulls two features representing similar samples closer to each other and pushes those of diverse samples far away from each other~\cite{wang2020understanding}, thus facilitating the DCNNs to learn various visual patterns without annotations. 
		Generally, without supervision, similar/positive samples of CL are obtained by augmenting the same image, and the diverse/negative samples can refer to different images. 
		To learn from unlabelled face images, existing CL-based methods \cite{zheng2022general,roy2021self,shu2022revisiting} have achieved effective self-supervised facial representation learning. 
		%如果上面没有参考文献，用下面这句话Naturally, CL methods could be utilized for unsupervised/self-supervised facial representation learning.

		However, despite progress, we found that directly utilizing CL-based methods still obtained sub-optimal performance due to the facial poses.  
		In particular, CL-based methods treat the augmented images from the same image as positive samples. In such a manner, the learned features are pose-invariant, which cannot recognize the variances of facial poses. Nevertheless, poses are one significant consideration for facial understanding~\cite{samanta2017role, adams2015decoupling}; for example, a person tends low their head when they feel sad.

		To tackle the above limitation of CL, we propose a Pose-disentangled Contrastive Learning (PCL) method, which disentangles the learning on pose-related features and pose-unrelated facial features for CL-based self-supervised facial representation learning. Fig.~\ref{fig:motivation} has shown an intuitive example of contrastive learning results.
		Specifically, Our method introduces two novel modules, \ie, a pose-disentangled decoder (PDD) and a pose-related contrastive learning scheme (see Fig.~\ref{fig:networks}). In the PDD, we first obtain the face-aware features from a backbone, such as ResNet  \cite{he2016deep,8707128}, Transformer~\cite{dosovitskiyimage, Chen_2022_CVPR}, and then disentangle pose-related features and pose-unrelated facial features from the face-aware features using two different subnets through facial reconstruction. In facial reconstruction, the combination of one pose-unrelated facial feature and one pose-related feature can reconstruct an image with the same content as the pose-unrelated facial feature and the same pose as the pose-related feature. Furthermore, an orthogonalizing regulation is designed to make the pose-related and pose-unrelated features more independent. 
		
		In the pose-related contrastive learning, instead of learning pose-invariant features by normal CL, we introduce two types of data augmentation for one face image, one containing pose augmentation and another only containing pose-unrelated augmentation. Therefore, image pairs generated by using pose augmentation contain different poses and serve as negative pairs, whereas image pairs generated from pose-unrelated augmentation contain the same pose as the original image and are treated as positive pairs. The pose-related CL is conducted to learn pose-related features, and face CL is used to learn pose-unrelated facial features. Therefore, our proposed pose-related CL can learn detailed pose information without disturbing the learning of pose-unrelated facial features in the images.

		In general, the major contributions of this paper are summarized as follows:
		\begin{enumerate}
			\item 
			{\lyy We propose a novel pose-disentangled contrastive learning framework, termed PCL, for learning unlabeled facial data. Our method introduces an effective mechanism that could disentangle pose features from facial features and enhance contrastive learning for pose-related facial representation learning.  
				\item We introduce a PDD using facial image reconstruction with a delicately designed orthogonalizing regulation to help effectively identify and separate the face-aware features obtained from the backbone into pose-related and pose-unrelated facial features. The PDD is easy-to-implement and efficient for head pose extraction. 
				\item We further propose a pose-related contrastive learning scheme for pose-related feature learning. Together with face contrastive learning on pose-unrelated facial features, we make both learning schemes cooperate with each other adaptively and obtain more effective learning performance on the face-aware features. 
				\item Our PCL can be well generalized to several downstream tasks, \eg, facial expression recognition (FER), AU detection, facial recognition and head pose estimation. Extensive experiments show the superiority of our PCL over existing SSL methods, accessing state-of-the-art performance on self-supervised facial representation learning.}
		\end{enumerate}
		\begin{figure*}[t]
			\begin{center}
				\includegraphics[width=0.75\linewidth]{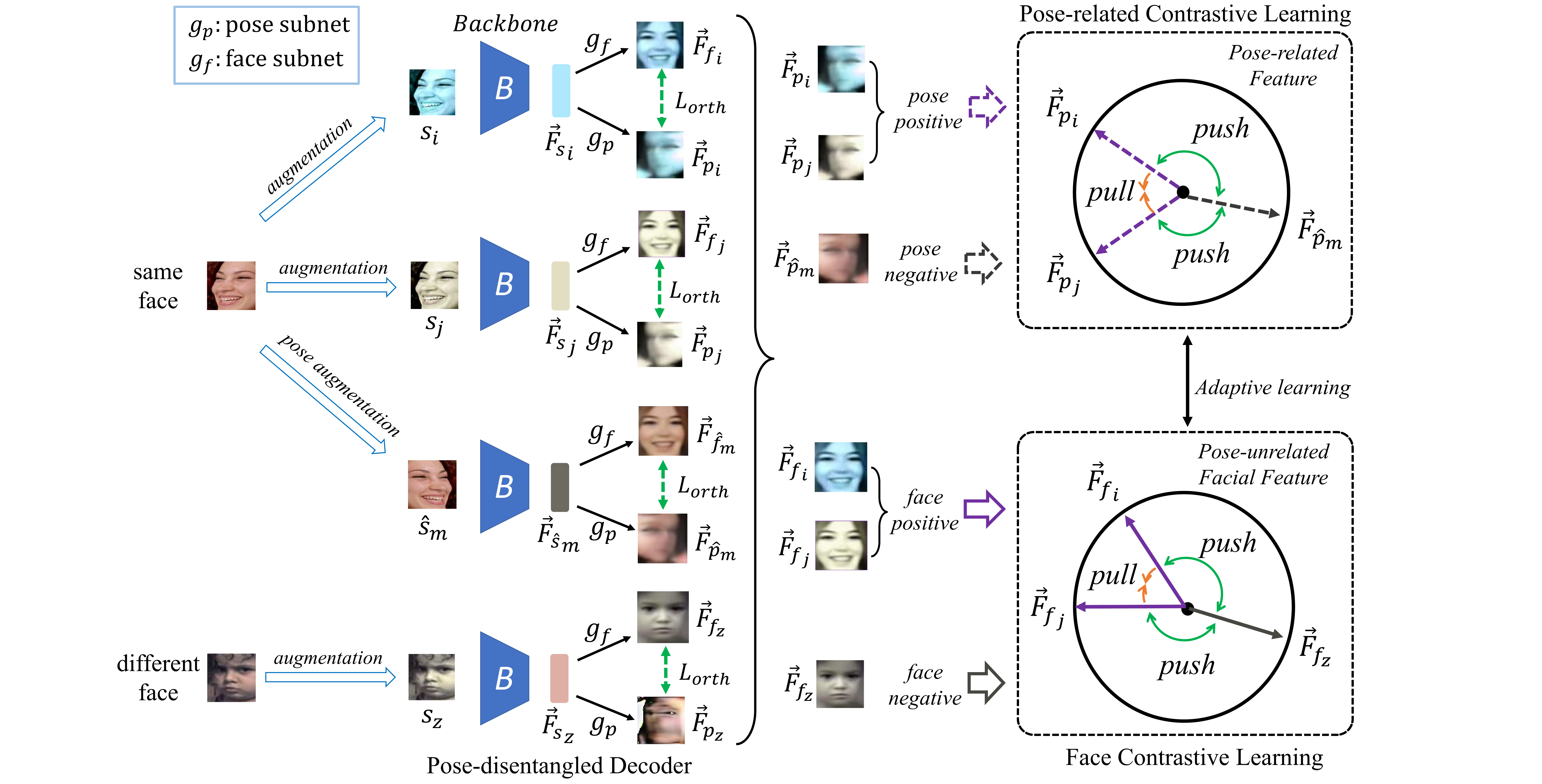}
			\end{center}
			\vspace{-0.3cm}
			\caption{
				The overview of PCL for self-supervised facial representation learning. 
				We first use a pose-disentangled decoder with an orthogonalizing regulation $L_{orth}$ to help extract pose-related features (\eg, $\vec{F}_{pi}$) and pose-unrelated facial features (\eg., $\vec{F}_{fi}$) from input augmented images (\eg, $s_i$), and then introduce pose-related contrastive learning and face contrastive learning schemes to further learn on the extracted features adaptively, resulting in more effective face-aware representation learning.	
			}
			\label{fig:networks}
			\vspace{-0.3cm}
		\end{figure*}
		
		\section{Related Work}
		
		\paragraph{Contrastive Learning}
		
		Contrastive learning (CL) has been widely used in self-supervised learning and has yielded significant results in many vision tasks~\cite{chen2020simple,he2020momentum,he2021masked,chen2021exploring,chen2021empirical,pan2021videomoco,xie2021detco,chen2020improved,zhu2021improving}.
		CL aims to map features of samples onto a unit hypersphere such that the feature distances of the positive sample pairs on the sphere are similar. In contrast, the feature distances of the randomly sampled negative sample pairs are pushed apart~\cite{wang2020understanding}. Recent breakthroughs in CL, such as MoCo~\cite{he2020momentum} and SimCLR~\cite{chen2020simple}, shed light on the potential of discriminative models for visual representation. 
		Thanks to a large number of negative samples, MoCo maintained a queue of negative samples to improve the capacity of CL~\cite{he2020momentum}. 
		Chen \textit{et al.}~\cite{chen2020simple} proposed a new self-supervised framework SimCLR to model the similarity of two images for learning visual representations without human supervision. 
		SimSiam is proposed for exploring simple siamese representation learning by maximizing the similarity between two augmentations of one image, subject to certain conditions for avoiding collapsing solutions \cite{chen2021exploring}.
		
		\paragraph{Self-supervised Facial Representation Learning}
		
		Self-supervised facial representation learning is important for many face-related applications, such as FER, face recognition, AU detection, etc.~\cite{chang2021learning,xue2021transfer,chang2006manifold,li2019self,wiles2018self,hu2018seqface,chang2022knowledge}. %Due to the difficulty of labeling face data for different face tasks, a lot of work has begun to focus on how to use large amounts of unlabeled data to learn self-supervised facial representation.
		Due to its capacity of learning on unlabelled data, an increasing number of research efforts are focusing on self-supervised face representation learning.
		FAb-Net used the motion changes between different frames of a video to learn facial motion features, and has achieved good results in FER~\cite{wiles2018self}. Li \textit{et al.} \cite{li2019self,li2020learning} proposed a Twin-Cycle Autoencoder that can disentangle the facial action-related movements and the head motion-related ones, obtaining good facial emotion representation for self-supervised AU detection~\cite{li2019self}. 
		FaceCycle decoupled facial expression and identity information via cyclic consistency learning to extract robust unsupervised facial representation, thus achieving good results in both FER and facial recognition~\cite{chang2021learning}. 
		{\lyy Zheng \textit{et al.} presented an study about the transferable
			visual models learned in a visual-linguistic manner on
			general facial representations \cite{zheng2022general}.}
		Roy \textit{et al.}~\cite{roy2021self} proposed a CL-MEx for pose-invariant expression representation by exploiting facial images captured from different angles. Shu \textit{et al.}~\cite{shu2022revisiting} used three sample mining strategies in CL to learn expression-related features. %still ignoring the effects of pose-related information. 
		Overall, most of the existing work is crafting facial representation learning for a single task, and general self-supervised facial representation learning remains an open research problem.

		\section{The Proposed Approach}
		The overview of our proposed pose-disentangled contrastive learning is presented in Fig.~\ref{fig:networks}. Our PCL mainly consists of two novel modules, \ie, a pose-disentangled decoder (PDD) and a pose-related contrastive learning scheme. Tacking a face image as input, the PDD of PCL first employs a backbone network like ResNet  \cite{he2016deep,8707128} to extract general facial features and then attaches two subnets to produce separate pose-related features and pose-unrelated facial features. To train the PDD properly, we reconstruct the face through the combination of the two types of features with an orthogonalizing regulation posed on the separated features for better disentanglement. Then, we introduce pose-related contrastive learning to train the pose-related features and use face contrastive learning scheme to learn pose-unrelated facial features. We make the two learning objectives cooperate with each other adaptively, obtaining more promising self-supervised facial representations. Our PCL method can fulfill the training of neural networks in an end-to-end manner. In the following sections, we will describe the details of PCL.
		
		% {\cz Our PCL mainly consists of two novel modules, \ie, a pose-disentangled decoder (PDD) and a pose-related contrastive learning scheme. Tacking a face image as input, our PCL method first applies the PDD to disentangle face information and extract pose-related features and pose-unrelated facial features. The PDD employs a backbone network like ResNet \cite{he2016deep} to extract general facial features and then attaches two subnets to produce separate pose-related features and pose-unrelated facial features with an orthogonalizing regulation posed on the separated features for better disentanglement. 
		% To train the PDD properly, we attempt to reconstruct pose-varied faces. 
		% Then, we introduce pose-related contrastive learning and face contrastive learning schemes to train the pose-related features and pose-unrelated facial features, respectively. 
		% We make the two learning objectives cooperate with each other adaptively, obtaining more promising self-supervised facial representations after pre-training. Our PCL method can fulfill the training of neural networks in an end-to-end manner. In the following sections, we will describe the details of PCL.
		% }

		\subsection{Pose-disentangled Decoder }
		Previous CL-based methods \cite{chen2020simple,he2020momentum} treat pose and other facial information uniformly, resulting in pose-invariant features that cannot recognize the details of poses. One possible solution is not using pose augmentation for training. However, such a manner would reduce the training data diversity and further reduce the performance. To conquer the above limitation of CL, we design a PDD to disentangle the pose-related and pose-unrelated facial representations from the face-aware features. Therefore, through the individual learning information from the pose-related and pose-unrelated facial features, the face-aware features could be used as a well facial representation that properly depicts the face, including the pose and other useful information.
		
		%The previous CL-based methods \cite{chen2020simple,he2020momentum} treat pose and other face information uniformly which leads to that  the network cannot effectively cognize the effects of pose characteristics. These methods either do not introduce pose augmentation, making the performance poor, or introduce pose augmentation to learn pose-invariant features. Neither of these methods achieves good results. To conquer the above limitation of CL, we design a PDD with  a delicately designed orthogonalizing regulation to disentangle the pose-related features and other pose-unrelated facial features from the face-aware features and do not affect each other’s training. 
		
		Nevertheless, identifying and separating the pose-related and pose-unrelated facial features is nontrivial. 
		To tackle this problem, the PDD assists the training of the two types of features through reconstructing faces: one image with a specific pose could be reconstructed through the combination of the pose-unrelated facial feature of the corresponding image and the pose-related feature of the given pose.

		The overall pipeline of PDD is presented in Fig.~\ref{fig:reconstruct}. {\lyy PDD consists of a shared backbone network, two separating subnet branches, and a shared reconstruction network.} 
		{\lyy In this paper, the backbone network is a shallow 16-layer residual network 
			for learning facial features from input face image.} Note that our PCL can marry with any other backbones, such as VGG and Transformer. Then, 
		two extra separating subnets attach to the {\lyy backbone} are used to separate the obtained features into the pose-related features and the {\lyy pose-unrelated facial features} (both features are 2048-dimensional features in practice), respectively. 
		Finally, {\lyy we employ a 6-layer blocks with each of an upsampling layer and convolutional layer as the reconstruction network to translate the combination of the pose-related features and pose-unrelated facial features into a reconstructed face. }

		\begin{figure}[t]
			\begin{center}
				\includegraphics[width=1.\linewidth]{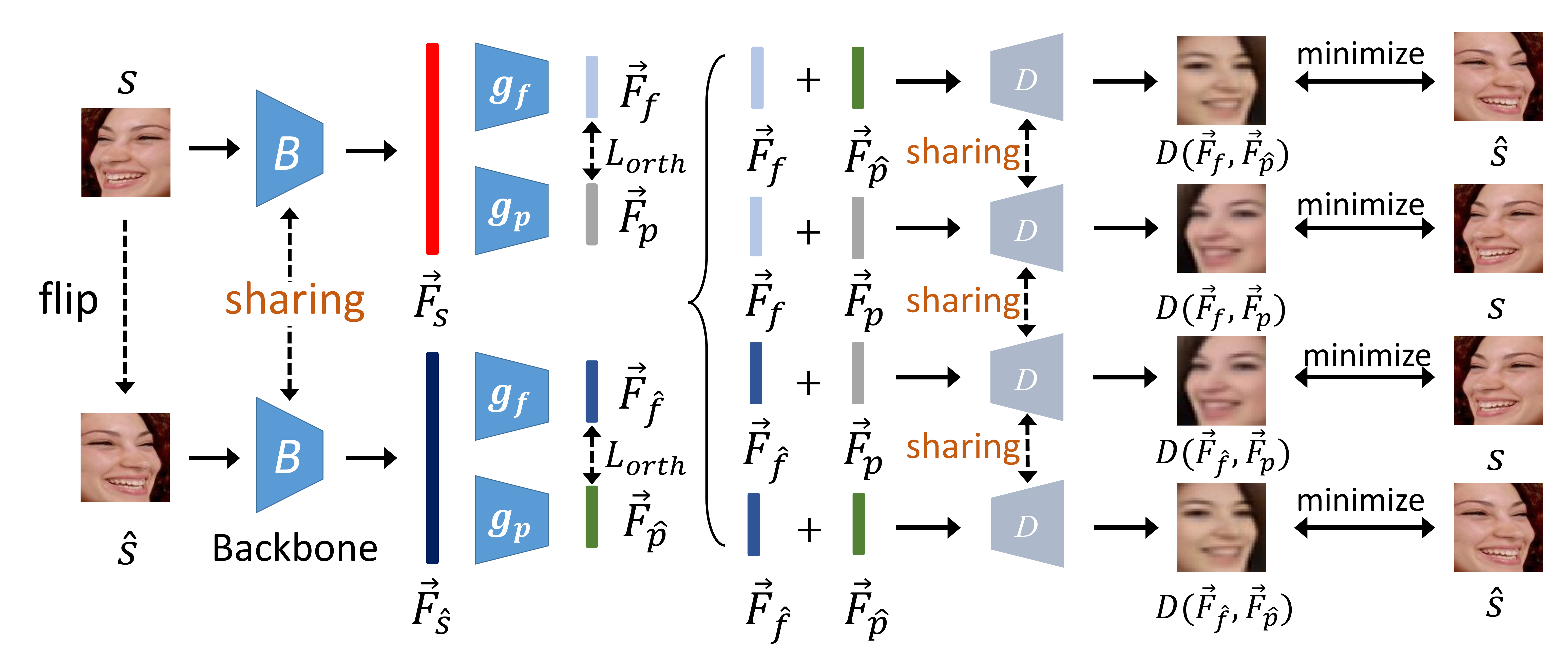}
			\end{center}
			\vspace{-0.3cm}
			\caption{
				The training pipeline of PDD. Given the face image $s$ and its pose-varied image $\hat {s}$ as input, we first use a backbone to encode facial features of input images, and then use two separating subnets, \ie, $g_{f}(\cdot)$ and $g_p(\cdot)$, to extract pose-unrelated and pose-related facial features, respectively. Finally, we employ a face reconstruction network $D$ to translate the extracted two types of features into reconstructed faces. Moreover, an orthogonalizing regulation $L_{orth}$ is used for training the PDD to make the separated features independent of each other.}
			\label{fig:reconstruct}
			\vspace{-0.3cm}
		\end{figure}
		
		Formally, given an input face image $s$, we represent its pose as $p$. {\cz We represent the same face with a different pose as $\hat{s}$ with its pose $\hat{p}$}. 
		In PDD, we use the backbone \textit{B} to encode face data into a facial feature $\vec{F}_s$ with the pose. 
		{\cz We would like to mention again that the final learned $\vec{F}_s$ is the self-supervised face-aware representation for the downstream task evaluation.} 
		Then, two separating branches, denoted as $g_p(\cdot)$ and $g_f(\cdot)$, are employed to extract the pose-related feature $\vec{F}_p$ and the {\lyy pose-unrelated facial feature} $\vec{F}_f$, respectively. 
		Meanwhile, using the same backbone and separating branches, we have corresponding features $\vec{F}_{\hat{f}}$ and $\vec{F}_{\hat{p}}$ for the pose-varied face $\hat{s}$. According to the goal of PDD, the $\vec{F}_f$ and $\vec{F}_{\hat{f}}$ are supposed to represent the same facial features, while the $\vec{F}_p$ and $\vec{F}_{\hat{p}}$ are supposed to describe different pose-related features. To achieve this, we introduce a reconstruction network $D$ to translate pose-related and {pose-unrelated facial} features into reconstructed faces that can be defined explicitly. As a result, we suppose the PDD should satisfy the following transformations:
		\begin{equation}
			\begin{aligned}
				D(\vec{F}_f, \vec{F}_p) = s,\ 
				D(\vec{F}_f, \vec{F}_{\hat{p}}) = \hat{s}, \\
				D(\vec{F}_{\hat{f}}, \vec{F}_{p}) = s,\    
				D(\vec{F}_{\hat{f}}, \vec{F}_{\hat{p}}) = \hat{s}.
			\end{aligned}.
			\label{eq:decode}
		\end{equation}
		The above goals indicate that the PDD should reconstruct the same face but different poses according to varied pose-related features. When the above transformations can be satisfied, we can then consider that the PDD {\cz tends to have} the ability to separate the pose-related feature $F_p$ from {\lyy pose-unrelated facial feature} $F_f$ properly. Otherwise, for example, if $\vec{F}_f$ still contains redundant feature about $p$, the $D(\vec{F}_f, \vec{F}_{\hat{p}})$ would not generate the $\hat{s}$ image appropriately {\cz and would tend to produce the $s$ instead.}

		{\cz To make PDD satisfy the above transformations}, the {\lyy disentangled} objective $L_{dis}$ of the PDD is:
		\begin{equation}
			\begin{split}
				L_{dis} &= ||s-D(\vec{F}_f, \vec{F}_p)||_1 + ||\hat{s}-D(\vec{F}_f, \vec{F}_{\hat{p}})||_1 \\  
				&+ ||s-D(\vec{F}_{\hat{f}}, \vec{F}_{p})||_1 + ||\hat{s}-D(\vec{F}_{\hat{f}}, \vec{F}_{\hat{p}})||_1, 
				\label{eq:P-DAE}
			\end{split}
		\end{equation}
		where $|| \cdot ||_1$ represents $l_1$-norm. {\lyy Additionally, we also try to use  GAN~\cite{goodfellow2014generative} instead of the $l_1$-norm; however, GAN can only make generated images approximate the real images but cannot guarantee the poses of the generated images. For more discussion, see supplemental material.}
		
		Moreover, to disentangle the extracted features more properly, an orthogonalizing regulation is further introduced to make the extracted features uncorrelated. {\lyy Therefore, we constrain that the $\vec{F}_{f}$ and $\vec{F}_{p}$ should be orthogonal to each other. To achieve this, inspired by \cite{hazarika_misa_2020,bousmalis2016domain,liu2017adversarial,ruder2018strong}, the orthogonalizing regulation $L_{orth}$ is defined as follows: %for disentangling pose-related features and pose-unrelated facial features, 
			\begin{equation}
				% 	L_{orth} =  \sum^{N}_{k=1}{||\vec{F}_{f_i} \cdot \vec{F}_{p_i}||^2_2 + ||\vec{F}_{f_j} \cdot \vec{F}_{p_j}||^2_2}.
				L_{orth} =  \frac{1}{N} (\sum^{N}_{i=1}{||\vec{F}_{f} \cdot \vec{F}_{p}||^2_2} + \sum^{N}_{i=1}{||\vec{F}_{\hat{f}} \cdot \vec{F}_{\hat{p}}||^2_2}).
			\end{equation}
			
			During learning, minimizing the $L_{orth}$ can help force the dot-products of pose-related and  pose-unrelated facial features to reach near zero, thus making them orthogonal to each other. Finally, we define the total optimization objective $L_{PDD}$ of the PDD, including the disentangled loss and orthogonalizing regulation as:}
		\begin{equation}
			% 	L_{orth} =  \sum^{N}_{k=1}{||\vec{F}_{f_i} \cdot \vec{F}_{p_i}||^2_2 + ||\vec{F}_{f_j} \cdot \vec{F}_{p_j}||^2_2}.
			L_{PDD} =  L_{orth} + L_{dis}. %+ \sum^{N}_{j=1}{||\vec{F}_{f_j} \cdot \vec{F}_{p_j}||^2_2}).
		\end{equation}
		
		\subsection{Pose-related Contrastive Learning}
		
		Normal CL tends to learn pose-invariant features. Therefore,
		{\lyy we further devise a Pose-related Contrastive Learning to enable effective self-supervised learning on pose information, suppressing the side effects of pose-invariant features. 
			Since it is unknown whether different faces have the same pose or not, it is difficult to construct pose positive and negative sample pairs well by directly using data augmentation in contrastive learning.
			To address this problem, unlike the normal CL that treats different face individuals as negative pairs, we propose a pose augmentation method for pose-related contrastive learning, \ie, for the same face image, we apply pose transformation and image transformation to it separately, then consider the same image pairs containing different (augmented) poses as negative pairs and the same image with containing the same (unaugmented) pose as positive pairs for contrastive learning on the pose. 
			Through this way, pose-related contrastive learning can focus on learning pose information without being influenced by images with the same pose as the negative samples. 
			
			Formally, for the input face image $s$, we use the specific pose augmentation (such as flipping and rotation) %{\wwb and stochastic data augmentation (such as random crop, color jitter, Gaussian blur and Sobel filtering)} 
			to generate $M$ negative samples $\hat{s}_M$, \ie, $\hat{s}_M=\{\hat{s}_{m}\}_{m=1}^M$, resulting in a negative pair ($\hat{s}_i$ and $\hat{s}_{m}$), while using a stochastic data augmentation (such as, random crop, color jitter, Gaussian blur, and Sobel filtering) to obtain a positive pair ($s_i$ and $s_j$), as shown in  Fig.~\ref{fig:networks}. Both the positive and negative pairs are passed through the PDD to extract the pose-related features as $\vec{F}_{p_i}, \vec{F}_{p_j}$ and $\vec{F}_{\hat{p}_m}$, respectively. Overall, the pose-related contrastive loss is written as:
			\begin{equation}
				\begin{aligned}
					L_{pose}(\vec{F}_{p_i},\vec{F}_{p_j},\vec{F}_{\hat{p}_m}) =  l_p(\vec{F}_{p_i},\vec{F}_{p_j})+l_p(\vec{F}_{p_j},\vec{F}_{p_i}), \\
					l_p(\vec{F}_{p_i},\vec{F}_{p_j}) = -log\frac{exp(\frac{sim(\vec{F}_{p_i}, \vec{F}_{p_j})}{\tau})}{\sum_{m=1}^{M} exp(\frac{sim(\vec{F}_{p_i}, \vec{F}_{{\hat{p}_m}})}{\tau})},\\
				\end{aligned}
			\end{equation}
			where $sim(\cdot)$ is the pairwise cosine similarity. $\tau$ denotes a temperature parameter. Through the pose-related contrastive learning, our PCL can learn more detailed pose information from facial images without disturbing the learning of pose-unrelated facial features.}
		%where $sim(\cdot)$ is the pairwise cosine similarity and $\tau$ is a temperature parameter in contrastive learning.

		\subsection{Overall Optimization Objectives}
		
		{\lyy 
			Together with pose-related contrastive learning on the pose-related features, we employ face contrastive learning on the pose-unrelated facial features, thus using two different subnetworks with different CL strategies to alleviate the side effects of pose information for learning face patterns.} 
		More specifically, we randomly sample a minibatch of $N$ face images, and use a stochastic data augmentation to transform any given input face image $s$, resulting in two correlated views of the same face as a positive pair $s_i$ and $s_j$. Secondly, each positive pair, \eg, $s_i$ and $s_j$ in Fig.~\ref{fig:networks}, are passed through the PDD to extract the {\lyy pose-unrelated facial features} $\vec{F}_{f_i} $ and $\vec{F}_{f_j}$, respectively. %Given a pair of positive $\vec{F}_{f_i} $  and $\vec{F}_{f_j}$, 
		The contrastive loss {\lyy on the face branch} is written as:
		\begin{equation}
			\begin{aligned}
				L_{face}(\vec{F}_{f_i},\vec{F}_{f_j}) =  l_f(\vec{F}_{f_i},\vec{F}_{f_j})+l_f(\vec{F}_{f_j},\vec{F}_{f_i}), \\
				l_f(\vec{F}_{f_i},\vec{F}_{f_j}) = -log\frac{exp(\frac{sim(\vec{F}_{f_i}, \vec{F}_{f_j})}{\tau})}{\sum_{z=1}^{2N} 1_{[i\ne z]}exp(\frac{sim(\vec{F}_{f_i}, \vec{F}_{f_z})}{\tau})},\\
			\end{aligned}
		\end{equation}
		where $\vec{F}_{f_z}$ is from negative pairs.%represents face features from different individuals.

		{\lyy Therefore, during training, our PCL has three {\cz major} objectives: the disentangled lose $L_{PDD}$ of PDD, the pose-related contrastive loss $ L_{pose}$ on the pose-related features, and the face contrastive loss $L_{face}$ on the pose-unrelated facial features. 
			The overall objective function $L$ of the PCL is the weighted sum of $L_{PDD}$, $L_{pose}$, and $L_{face}$. Mathematically, the total loss $L$ can be written as:
			\begin{equation}
				L= L_{PDD} +  \alpha_{pose} \cdot L_{pose} + \alpha_{face} \cdot L_{face},
			\end{equation}
			where $\alpha_{pose}$ and $\alpha_{face}$ are two dynamic weights to adaptively balance the pose and face learning objectives in the multi-task learning manner according to their contributions to facial representations. 
			We employ the Dynamic Weight Average (DWA) \cite{liu2019end} to obtain the $ \alpha_{pose}$ and $\alpha_{face}$ during training. More details of dynamic weight learning can be seen in the supplemental material.
			We also show in the experiments that adding the dynamic weight learning improves performance (see Table~\ref{table:components}), demonstrating the usefulness of adaptive cooperation of two CL schemes.
	}}

	\section{Experiments}
	In this section, we verified the effectiveness of our proposed PCL by answering two questions:
	
	Q1: does our facial representation perform well and has generalizability? (Refer to section 4.2)
	
	Q2: whether the improvements come from the contributions we proposed in this paper? (Refer to section 4.3)
	
	We further visualized the contents of learned features to demonstrate the reasonability of PCL. (Refer to section 4.4)
	
	%Extensive experiments show that the acquired self-supervised facial representation generalizes to a range of face tasks.
	
	\subsection{Experimental Settings}
	\paragraph{Datasets}
	The proposed PCL was trained on the combination of VoxCeleb1~\cite{Nagrani17} and VoxCeleb2~\cite{Chung18b} datasets without any annotations. 
	The VoxCeleb1 and VoxCeleb2 have 299,085 video clips of around 7,000 speakers. We extracted the video frames at 6 fps, cropped to faces shown in the center of frames and then resized to the resolution of 64 $\times$ 64 for training~\cite{chang2021learning}.
	%Video frames were extracted at 6 fps, cropped to faces shown in the center of frames, and then resized to the resolution of 64 $\times$ 64 for training~\cite{chang2021learning}. 
	
	\textit{{For FER evaluation}}, we used two widely-used FER datasets, \ie, FER-2013~\cite{goodfellow2015challenges} and RAF-DB~\cite{li2017reliable}. The FER-2013 consists of 28,709 training and 3,589 testing images. 
	We followed the experimental setup as~\cite{chang2021learning} to particularly use the basic emotion subset of RAF-DB with 12,271 training and 3,068 testing images. 
	
	\textit{{For facial recognition evaluation}}, we adopt two in-the-wild facial identity datasets, \ie, LFW~\cite{huang2008labeled} and CPLFW~\cite{zheng2018cross}. The LFW consists of 13,233 face images from 5,749 identities and has 6,000 face pairs for evaluating identity verification. 
	The CPLFW dataset includes 3,000 positive face pairs with pose differences to add pose variation to intra-class variance.
	All reported results were averaged across the 10 folds.
	
	\textit{{For facial AU detection}}, we evaluated our method on the DISFA~\cite{mavadati2013disfa} dataset with 26 participants. 
	%BP4D contains 41 participants (23 females and 18 males). There are about 146000 frames with available AU labels. 
	%DISFA consists of 26 participants. 
	The AUs are labeled with intensities from 0 to 5. The frames with intensities greater than 1 are considered positive, while others are treated as negative. In total, we obtained about 130,000 AU-labelled frames and followed the experimental setup of ~\cite{li2019self} to conduct a 3-fold cross-validation. %to split the dataset into 3 folds based on subject IDs and 

	\textit{{For head pose estimation}}, we adopt two widely-used tasks, \ie, pose regression (pretrained on 300W-LP~\cite{sagonas2013300} and evaluated on AFLW2000~\cite{zhu2016face}) and pose classification (on BU-3DFE~\cite{yin20063d}). The 300W-LP contains 122,450 images and AFLW2000 contains 2000 images. We followed the experimental setup as~\cite{LIU2021195} to use BU-3DFE with 14,112 images as training datasets and 6,264 images as validation datasets.
	%we pretrained on 300W-LP~\cite{sagonas2013300} and evaluated on AFLW2000~\cite{zhu2016face}. The 300W-LP contains 122,450 images and AFLW2000 contains 2000 images. 
	%We perform evaluations on BU-3DFE~\cite{yin20063d} under nine pan angles. We followed the experimental setup as~\cite{LIU2021195} to use BU-3DFE with 14,112 images as training datasets and 6,264 images as validation datasets.
	
	%\paragraph{Implementation Details}
	\noindent{\bf Implementation Details}
	Our proposed model was implemented based on the PyTorch framework and trained with the Adam optimizer ($\beta_1=0.9$, and $\beta_2=0.999$) for 1000 epochs. 
	The batch size and initial learning rate are set to 256 and 0.0001, respectively. The learning rate is decreased by cosine annealing. The temperature parameter $\tau$ is set to 0.07.
	{\lyy The baseline SimCLR~\cite{chen2020simple} used the data augmentation (such as random crop, color jitter, Gaussian blur, and Sobel filtering) and negative interpolation~\cite{zhu2021improving} for training.}
	
	Referring to ~\cite{chang2021learning} and ~\cite{wiles2018self}, the backbone of our model is a simple 16-layer CNN, and the reconstruction network is a simple 6-layer block with each of an upsampling layer and a convolutional layer.
	The $g_f(\cdot)$ and $g_p(\cdot)$ are convolutional subnets with the same architecture. We will give the detailed network structure in the supplementary material.
	
	{\lyy In addition, 
		we explored different choices of varying the pose $p$ for training PDD like flipping and rotation. However, the experimental results demonstrate that flipping $p$ is the most effective way to help PDD learn to identify and separate pose from facial representations (0.43\% improvement over adding rotation and translation). To trade off between efficiency and accuracy, we used pose flipping for training PDD in this study. %$\hat{p}$ to only represent the flipped $p$ in PDD.} 

		%\paragraph{Evaluation Protocols}
		\noindent{\bf Evaluation Protocols}
		We followed the widely used linear evaluation protocol in SSL~\cite{chen2020simple,he2020momentum,he2021masked,chen2021exploring,grill2020bootstrap,chen2021empirical,chang2021learning,datta2018unsupervised,li2019self} to verify our method. 
		The linear classifier is a simple linear fully-connected layer, and is trained with the frozen self-supervised {\lyy face-aware representation} $\vec{F}_s$ from the backbone $B$ for 300 epochs.
		
		Following ~\cite{chang2021learning,datta2018unsupervised,li2019self}, we resized the images to the size $100 \times 100$, $128 \times 128$,  $256 \times 256$ and $256 \times 256$ respectively, for FER, face recognition, AU detection and pose-related downstream tasks. 
		
		\subsection{Performance Comparison for Q1}
		\subsubsection{Evaluation for Facial Expression Recognition}
		
		Given the trained model, we investigated the learned $\vec{F}_s$ by evaluating the performance of its applications on FER. The quantitative results shown in Table~\ref{table:exp} demonstrate that our proposed method is able to provide superior performance with respect to other methods. Compared to the SimCLR~\cite{chen2020simple}, the proposed PCL improves the accuracy by over 7.3\% and 3.41\%, respectively. These results suggest that our PCL can be used as a pretext task to learn an effective self-supervised facial representation with rich expression information for the FER task.
		\begin{table}[htp]
			\begin{center}
				\caption{Evaluation of the FER task on the FER-2013 and RAF-DB datasets. (Note: the highest results of self-supervised methods are highlighted in bold, and * indicates the results reproduced by authors.)}
				\label{table:exp}
				\resizebox{0.75\linewidth}{!}{
					\begin{tabular}{lcc}
						\hline
						\multicolumn{1}{l|}{}              & \multicolumn{1}{c|}{FER-2013}       & RAF-DB         \\ \hline
						\multicolumn{1}{l|}{Method}        & \multicolumn{1}{c|}{Accuracy(\%)}   & Accuracy(\%)   \\ \hline
						\multicolumn{3}{l}{Fully supervised}                                                      \\ \hline
						\multicolumn{1}{l|}{FSN~\cite{zhao2018feature}}           & \multicolumn{1}{c|}{67.60}          & 81.10          \\
						\multicolumn{1}{l|}{ALT~\cite{florea2019annealed}}           & \multicolumn{1}{c|}{69.85}          & 84.50          \\ \hline
						\multicolumn{3}{l}{Self-supervised (linear evaluation)}                                        \\ \hline
						\multicolumn{1}{l|}{LBP~\cite{ojala2002multiresolution}}           & \multicolumn{1}{c|}{37.89}          & 52.17          \\
						\multicolumn{1}{l|}{HoG~\cite{dalal2005histograms}}           & \multicolumn{1}{c|}{45.47}          & 63.53          \\
						\multicolumn{1}{l|}{FAb-Net~\cite{wiles2018self}}       & \multicolumn{1}{c|}{46.98}          & 66.72          \\
						\multicolumn{1}{l|}{TCAE~\cite{li2019self}}          & \multicolumn{1}{c|}{45.05}          & 65.32          \\
						\multicolumn{1}{l|}{BMVC'20~\cite{lu2020self}}       & \multicolumn{1}{c|}{47.61}          & 58.86          \\
						\multicolumn{1}{l|}{MoCo~\cite{he2020momentum}}          & \multicolumn{1}{c|}{47.24}          & 68.32          \\
						\multicolumn{1}{l|}{FaceCycle~\cite{chang2021learning}}     & \multicolumn{1}{c|}{48.76}          & 71.01          \\
						\multicolumn{1}{l|}{SimCLR~\cite{chen2020simple}*}        & \multicolumn{1}{c|}{49.51}          & 71.06          \\
						\multicolumn{1}{l|}{\textbf{Ours}} & \multicolumn{1}{c|}{\textbf{56.81}} & \textbf{74.47} \\ \hline
					\end{tabular}
				}
				
			\end{center}
			\vspace{-0.3cm}
		\end{table}
		
		\subsubsection{Evaluation for Facial Recognition}
		
		For the facial recognition task, our learned self-supervised face-aware features also outperform other self-supervised-based facial representations. 
		As shown in Table~\ref{table:identity}, our PCL achieved the best accuracy of 79.72\% and 64.61\%  on LFW and CPLFW, respectively, which are 3.75\% and 1.26\% better than the results of the state-of-the-art method. The improvements suggest that our PCL can be used as an effective pretext task for facial identity recognition.
		\begin{table}[]
			\begin{center}
				\caption{Evaluation of facial recognition on the LFW and CPLFW datasets. (Note: the highest results of self-supervised methods are highlighted in bold, and * indicates the results reproduced by authors.)}
				\label{table:identity}
				\resizebox{0.75\linewidth}{!}{
					\begin{tabular}{lcc}
						\hline
						\multicolumn{1}{l|}{}                 & \multicolumn{1}{c|}{LFW}            & CPLFW                            \\ \hline
						\multicolumn{1}{l|}{Method}           & \multicolumn{1}{l|}{Accuracy(\%)}   & \multicolumn{1}{l}{Accuracy(\%)} \\ \hline
						\multicolumn{3}{l}{Fully supervised}                                 \\ \hline
						\multicolumn{1}{l|}{VGG-Face~\cite{parkhi2015deep}}         & \multicolumn{1}{c|}{98.95}          & 84.00                            \\
						\multicolumn{1}{l|}{SphereFace~\cite{liu2017sphereface}}       & \multicolumn{1}{c|}{99.42}          & 81.40                            \\
						\multicolumn{1}{l|}{ArcFace~\cite{deng2019arcface}}          & \multicolumn{1}{c|}{99.53}          & 92.08                            \\ \hline
						\multicolumn{3}{l}{Self-supervised (Linear evaluation)}       \\ \hline
						%	\multicolumn{1}{l|}{VGG~\cite{datta2018unsupervised}*}              & \multicolumn{1}{c|}{71.48}          & -              %                  \\
						\multicolumn{1}{l|}{LBP~\cite{ojala2002multiresolution}}              & \multicolumn{1}{c|}{72.44}          & -                            \\
						% 		\multicolumn{1}{l|}{HoG~\cite{dalal2005histograms}*}              & \multicolumn{1}{c|}{62.73}          & 51.73                            \\
						\multicolumn{1}{l|}{VGG~\cite{datta2018unsupervised}}              & \multicolumn{1}{c|}{72.20}          & -                                \\
						\multicolumn{1}{l|}{MoCo~\cite{he2020momentum}*}             & \multicolumn{1}{c|}{65.88}          & 57.82                            \\
						\multicolumn{1}{l|}{SimCLR~\cite{chen2020simple}*}             & \multicolumn{1}{c|}{75.97}          & 62.25                            \\
						%	\multicolumn{1}{l|}{FaceCycle~\cite{chang2021learning}*}        & \multicolumn{1}{c|}{73.72}          & 58.52            %                \\
						%		\multicolumn{1}{l|}{Ours}             & \multicolumn{1}{c|}{66.5}           & 50.51                       %         \\ \hline
						%		Linear evaluation                     & \multicolumn{1}{l}{}                & \multicolumn{1}{l}{}          %   \\ \hline
						\multicolumn{1}{l|}{FaceCycle~\cite{chang2021learning}*}        & \multicolumn{1}{c|}{74.12}          & 63.35                       \\
						\multicolumn{1}{l|}{\textbf{Ours}}    & \multicolumn{1}{c|}{\textbf{79.72}} & \textbf{64.61}                       \\ \hline
					\end{tabular}
				}
			\end{center}
			\vspace{-0.3cm}
		\end{table}

		%\vspace{-0.4cm}
		\subsubsection{Evaluation for Facial AU Detection}
		Facial AU detection estimates whether each AU in the face image or video is activated. 
		% After the self-supervised training process, we obtained the encoder $E$ for AU detection. 
		We followed the~\cite{li2019self} and used a binary cross-entropy loss to train a linear classifier for AU detection.
		Table~\ref{table:disfa} reports the comparison of our PCL and the state-of-the-art self-supervised methods, as well as the full supervised methods. 
		We evaluated not only the same backbone approaches as ours but also deeper backbone approaches. The results show that our method still has a clear advantage.
		As shown in Table~\ref{table:disfa}, our method outperforms other self-supervised methods in the average F1 score. Thanks to disentangled facial features, the learned facial representation can better reflect  facial actions. 
		In addition, our PCL has reached the fully supervised level, and the average $F1$ has exceeded the full supervised DRML~\cite{zhao2016deep} by 28.1 and the EAC-Net~\cite{li2017eac} by 6.3, respectively.
		\begin{table}[htpb]
			\begin{center}
				\caption{Evaluation of facial AU detection on the DISFA dataset. We use $F1$ score for the evaluation. (Note: the highest results of self-supervised methods are highlighted in bold, and * indicates the results reproduced by authors.)}
				\label{table:disfa}
				\resizebox{1.\linewidth}{!}{
					\begin{tabular}{c|l|llllllll|l}
						\hline
						\multicolumn{1}{l|}{}             & Methods/AU    & 1             & 2             & 4             & 6             & 9             & 12            & 25            & 26            & ave           \\ \hline
						\multirow{3}{*}{Supervised}       & DRML~\cite{zhao2016deep}          & 17.3          & 17.7          & 37.4          & 29.0          & 10.7          & 37.7          & 38.5          & 20.1          & 26.7          \\
						& EAC-Net~\cite{li2017eac}       & 41.5          & 26.4          & 66.4         & 50.7          & 80.5          & 89.3          & 88.9          & 15.6          & 48.5          \\
						& JAA-Net~\cite{shao2018deep}       & 43.7          & 46.2          & 56.0          & 41.4          & 44.7          & 69.6          & 88.3          & 58.4          & 56.0          \\ \hline
						\multirow{6}{*}{Self-superivised} & SplitBrain~\cite{zhang2017split}    & 13.1          & 10.6          & 35.7          & 40.2          & 30.2          & 57.5          & 77.4          & 40.3          & 38.1          \\
						& DeformAE~\cite{shu2018deforming}      & 17.6          & 12.3          & 46.7          & 43.5          & 26.0          & 62.7          & 64.8          & \textbf{47.6}          & 40.1          \\
						& Fab-Net~\cite{wiles2018self}       & 15.5          & 16.2          & 43.2          & \textbf{50.4}          & 23.2          & 69.6          & 72.4          & 42.4          & 41.6          \\
						& TCAE~\cite{li2019self}          & 15.1          & 16.2          & 50.5          & 48.7          & 23.3          & 72.1          & 72.4          & 42.4          & 45.0          \\
						& TCAE~\cite{li2019self}*          & 10.5          & 13.3         & 20.9         & 18.8          & 7.5          & 44.7         & 57.8          & 9.9          & 22.9          \\
						& FaceCycle~\cite{chang2021learning}*     & 26.4          & 10.2          & 37.3          & 21.5          & 25.0          & 71.8          & 84.2         & 34.7          & 38.9          \\
						& SimCLR~\cite{chen2020simple}*     & 40.5          & 46.9          & 53.8          & 33.5          & 24.9          & 74.7          & 85.0         & 35.6          & 49.4          \\
						& \textbf{Ours} & \textbf{53.8} & \textbf{44.9} & \textbf{58.1} & 37.2 & \textbf{53.2} & \textbf{73.1} & \textbf{86.5} & 31.3 & \textbf{54.8} \\ \hline
					\end{tabular}
				}
			\end{center}
			\vspace{-0.3cm}
		\end{table}
		
		\subsubsection{Evaluation for Head Pose Estimation}
		For the head pose estimation, we tested our PCL on pose regression (trained on 300W-LP and evaluated on AFLW2000) and pose classification (on BU-3DFE). We compare with different SSL methods in Table.~\ref{table:head_pose_regression}. Our PCL achieve 12.36 in terms of mean absolute error (MAE) on AFLW2000 and the best accuracy of 98.95\% on BU-3DFE, which outperforms the self-supervised state-of-the-art methods.
		\begin{table}[h]
			%\vspace{-0.3cm}
			\centering
			\caption{Evaluation on head pose estimation. ($\downarrow$ represents the smaller is better. $\uparrow$ represents the larger is better.)}
			%\vspace{-0.2cm}
			\label{table:head_pose_regression}
			\resizebox{0.95\linewidth}{!}{
				\begin{tabular}{c|cccl|c}
					\hline
					& \multicolumn{4}{c|}{AFLW2000 (pretrained on 300W-LP)} & BU-3DFE \\
					& Yaw$\downarrow$   & Pitch$\downarrow$ & Roll$\downarrow$  & MAE$\downarrow$   &    Accuracy (\%)$\uparrow$                      \\ \hline
					FaceCycle~\cite{chang2021learning} & 11.70 & \textbf{12.76} & 12.94 & 12.47 & 98.82                    \\
					MoCo~\cite{he2020momentum}    &    28.49   &    16.29   &   15.55    &    20.11   & 75.33                    \\
					SimCLR~\cite{chen2020simple} & 11.20 & 19.86 & 12.08 & 14.38 & 98.85 \\
					Ours      &   \textbf{9.86}    &   16.59    &   \textbf{10.62}    &   \textbf{12.36}    & \textbf{98.95}                    \\ \hline
				\end{tabular}
			}
			\vspace{-0.3cm}
		\end{table}
		
		\subsection{Ablation Study and Analysis for Q2}
		
		%\subsubsection{Effect of Different Modules}
		\begin{table}[t]
			\begin{center}
				\caption{Ablation study of the proposed PCL. Impact of integrating different components ( \ie, PDD and pose-related contrastive learning $L_{pose}$) into the baseline (SimCLR) on the RAF-DB dataset.}
				\label{table:components}
				\resizebox{0.95\linewidth}{!}{
					\begin{tabular}{ccc|cc|c}
						\hline
						\multirow{2}{*}{Baseline (SimCLR)} & \multicolumn{2}{c|}{PDD} & \multicolumn{2}{c|}{Contrastive learning} & \multirow{2}{*}{FER} \\ \cline{2-5}
						& $L_{dis}$          & $L_{orth}$     & $L_{pose}$     & Dynamic weighting      &                      \\ \hline
						\checkmark                  &            &            &      &                 &      71.06               \\
						\checkmark             &      \checkmark      &            &             &     &        71.47                    \\
						\checkmark      &    \checkmark         &     \checkmark        &            &      &        72.39                  \\
						\checkmark         &    \checkmark       &     \checkmark        &     \checkmark         &       &          73.73                \\
						\checkmark        &      \checkmark       &        \checkmark     &    \checkmark        &    \checkmark  &       \textbf{74.47}                  \\
						\hline
					\end{tabular}
				}
			\end{center}
			\vspace{-0.3cm}
		\end{table}

		\noindent{\bf Effect of Different Modules}
		%\paragraph{Effect of Different Modules}
		To better understand the role of each module in our PCL, Table~\ref{table:components} presents the ablation results of the gradual addition of different components into the baseline ({\lyy SimCLR w/o pose augmentation}) for FER on the RAF-DB dataset. The baseline achieved a FER accuracy of 71.06\%. Compared with the baseline, separating the pose-related features from face-aware features slightly improved the performance by 0.41\%. The further addition of $L_{orth}$ improved the FER accuracy to 72.39\%. We emphasized that this is the result of using  two normal contrastiveing learning schemes on the two  features separated by PDD.
		{\lyy A significant improvement of 1.34\% was obtained after adding the pose-related contrast learning $L_{pose}$, verifying that pose-related face information can help improve CL-based self-supervised facial representation performance. Additionally, the dynamic weighting achieved the best accuracy of  74.47\%.}

		\begin{table}[h]
			\caption{The effects of poses on SimCLR and our PCL (w/o Dynamic weighting).}
			%\vspace{-0.1cm}
			\label{table:simclr_flip}
			\resizebox{1\linewidth}{!}{
				\begin{tabular}{c|cccc}
					\hline
					Tasks         &     SimCLR w/o pose & SimCLR w/ pose & PCL w/o pose & PCL w/ pose \\ \hline
					FER(RAF-DB)                      & 71.06          &  73.17    & 73.24  &  \textbf{73.73}    \\
					Pose estimation(BU-3DFE)         & 98.93            &   98.85      & 98.40 &  \textbf{98.95}    \\ \hline
				\end{tabular}
			}
		\end{table}
		\noindent{\bf Effect of Poses on Contrastive Learning} 
		%\paragraph{Effect of Poses on Contrastive Learning}
		In order to further discuss  the pose-invariant face features learned by SimCLR~\cite{chen2020simple} and the pose-related face-aware features learned by our PCL,  Table~\ref{table:simclr_flip} shows the comparison of  SimCLR with and without pose augmentation, as well as our PCL with and without pose-related contrastive loss $L_{pose}$, respectively, on the RAF-DB dataset. SimCLR w/ pose achieved a relative accuracy increase of 2.97\% to SimCLR w/o pose on FER, while a relative decrease in pose estimation (about 0.09\%). The result demonstrates that learning pose-invariant features can help improve  CL performance. 
		
		In addition, PCL w/ pose achieved satisfied improvement in both FER (relative increase of 3.76\%) and pose estimation (an increase of 0.02\%), due to effectively exploring pose-unrelated facial and pose-related features. However, PCL w/o pose can not learn pose-related information, resulting in a slight decrease in both FER (decrease 0.49\%) and pose estimation (decrease 0.55\%). The experiment result shows that poses are one significant consideration for facial understanding.

		\begin{table}[]
			\begin{center}
				\caption{Linear evaluation with different face features. $\vec{F}_f+\vec{F}_p$ means to add the pose-related feature $\vec{F}_p$ with the pose-unrelated facial feature $\vec{F}_f$, and $\vec{F}_s$ represents the face-aware feature.}
				\label{table:projection}
				\resizebox{0.8\linewidth}{!}{
					\begin{tabular}{ccccc}
						\hline
						Different features & RAF-DB & LFW & DISFA \\ \hline
						$\vec{F}_f$ & 73.04      & 78.55   & 54.30     \\
						$\vec{F}_p$ &   65.71         &  62.55   &    34.17   \\
						$\vec{F}_s$ & \textbf{74.47}      & \textbf{79.72}   & 54.78     \\
						$\vec{F}_f + \vec{F}_p$ & 73.53     & 79.10  & \textbf{56.26}    \\ \hline
					\end{tabular}
				}
			\end{center}
		\end{table}
		
		\begin{figure}[t]
			\begin{center}
				\includegraphics[width=0.95\linewidth]{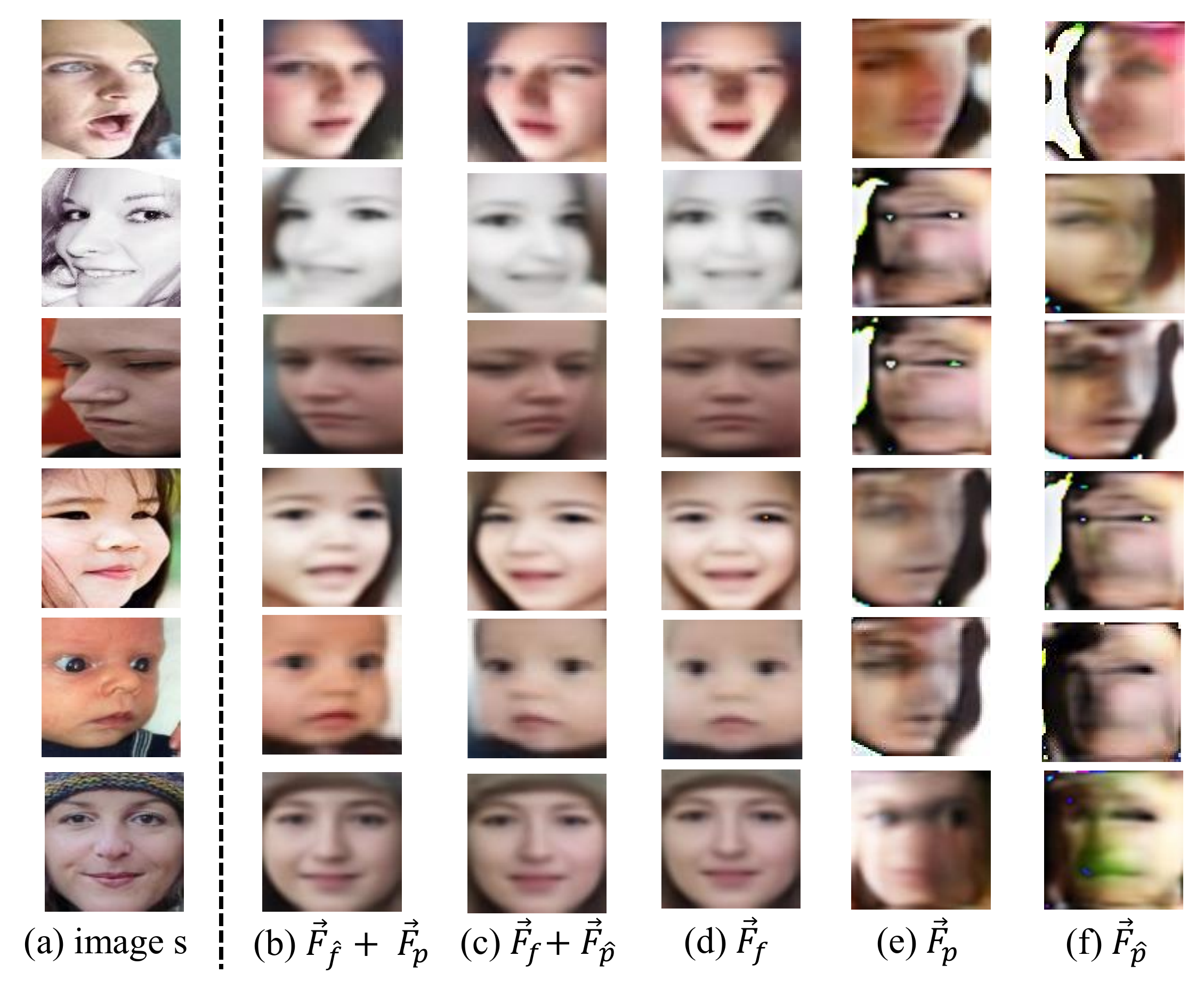}
			\end{center}
			%\vspace{-0.3cm}
			\caption{
				The reconstructed faces with disentangled pose-unrelated facial and pose-related features. (a)  Source image $s$, (b)-(f) the reconstructed faces with different features. $\vec{F}_f$:  pose-unrelated facial feature from $s$, $\vec{F}_p$:  pose-related feature from $s$, $\vec{F}_{\hat{f}}$: pose-unrelated facial feature from pose-flipped $\hat{s}$, $\vec{F}_{\hat{p}}$: pose-related feature from pose-flipped $\hat{s}$. } 
			\label{fig:recon}
			\vspace{-0.3cm}
		\end{figure}
		%\begin{figure}[t]
		%	\begin{center}
		%\includegraphics[width=0.6\linewidth]{fig/tsne_pose_related.png}
		%	\end{center}
		%	\vspace{-0.3cm}
		%	\caption{
		%		The pose-related and pose-unrelated facial features in 2D space by t-SNE visualization} 
		%	\label{fig:tsne}
		%	\vspace{-0.5cm}
		%\end{figure}
		%We study the importance of including the facial extractor $g_f(\cdot)$ in linear evaluation (after the encoder $E$, before the linear classifier). For a fair comparison, the facial images are rescaled to the size $64 \times 64$. 
		%\subsubsection{Comparison of Different Learned Features}
		\noindent{\bf Comparison of Different Learned Features} 
		Table~\ref{table:projection} shows a linear evaluation with different facial features extracted from the backbone and the followed two subnets, \ie, $\vec{F}_s$ extracted from the backbone $B$,  $\vec{F}_f$ extracted from $g_f(\cdot)$, and  $\vec{F}_p$ extracted from $g_p(\cdot)$, in our PCL. 
		For a fair comparison, the facial images were rescaled to the same size, and all the features were normalized to the same dimension in each case. The {\lyy face-aware features} $\vec{F}_s$ extracted from the backbone $B$ achieved the best performance for FER and facial recognition tasks, %on RAF-DB and LFW, 
		respectively. Compared with single $\vec{F}_f$, We added the pose-related features $\vec{F}_p$ with the pose-unrelated facial features $\vec{F}_f$ and gained improvement on three tasks by 0.49\%, 0.55\%, and 1.96\%, respectively. {\lyy The result demonstrates that pose-related information can be complementary to face information for achieving more effective face-aware representation in CL.}

		\subsection{Visualization}
		
		Fig.~\ref{fig:recon} visualizes the reconstructed faces with the pose-related and {\lyy pose-unrelated facial features} disentangled by our method.
		%We adopted RAF-DB testing images for visualizing the results of the disentanglement process. 
		As shown in Fig.~\ref{fig:recon}(b) and (c), our PCL successfully reconstructed the same faces but different poses according to varied pose-related features and the same pose-unrelated facial features, \ie $\vec{F}_{\hat{f}}+\vec{F}_p$ and $\vec{F}_f+\vec{F}_{\hat{p}}$, which shows the capability in separating pose-related features. 
		Fig.~\ref{fig:recon}(d) shows the reconstructed frontal faces with the pose-unrelated facial features $\vec{F}_f$ from the image $s$, 
		which demonstrates that our PCL is able to effectively disentangle the {\lyy facial features without poses}.
		Additionally, as shown in Fig.~\ref{fig:recon}(e) and (f), we just used pose-related features from the image $s$ and its pose-flipped image $\hat{s}$, \ie $\vec{F}_{p}$ and $\vec{F}_{\hat{p}}$. Obviously, the generated images only include varied pose information with few face patterns. 
		
		%Fig.~\ref{fig:tsne} visualized the pose-related features $\vec{F}_p$ and {\lyy pose-unrelated facial features $\vec{F}_f$} in a 2D feature space by using the t-SNE~\cite{van2008visualizing} on RAF-DB, demonstrating that our method can effectively separate pose-related features from pose-unrelated facial features.

		\section{Conclusions}
		
		In this paper, a novel pose-disentangled contrastive learning (PCL) is proposed for general self-supervised {\lyy facial representation learning}. PCL introduces two novel modules, \ie, a pose-disentangled decoder (PDD) and a pose-related contrastive learning scheme. First, the PDD with a designed orthogonalizing regulation learns to disentangle pose-related features from face-aware features, thus obtaining pose-related and other pose-unrelated facial features independent of each other. Then, together with face contrastive learning on pose-unrelated facial features, we further
		propose a pose-related contrastive learning scheme on pose-related features. Both two learning schemes cooperate with each other adaptively for more effective self-supervised facial representation learning performance.
		With the two components, the proposed PCL achieved a vastly improved performance on four downstream face tasks, ( \ie, facial expression recognition, facial recognition, facial AU detection and head pose estimation).
		Extensive experiments demonstrate that PCL is superior to other state-of-the-art self-supervised methods, obtaining strong robust self-supervised facial representation. 
		{\lyy In the future, we will continue to discuss the effects of other face-related attributions, such as ages, makeup and occlusion.} We believe the proposed approach can be well extended to decouple other relevant information for more robust self-supervised and unsupervised facial representation.
		
		%\noindent{\bf Acknowledgments.}
		\paragraph{Acknowledgments.}
		This work was partially supported by the National Natural Science Foundation of China grant (Grant No. 62076227, 62002090) and Wuhan Applied Fundamental Frontier Project under Grant (No. 2020010601012166). Dr Zhe Chen is supported by Australian Research Council Project IH-180100002. We are grateful to the AC panel and reviewers for their comments.
		
		\clearpage
		\bibliographystyle{ieee_fullname}
		\bibliography{egbib}
	\end{document}